\documentclass[letterpaper]{article} 
\usepackage{aaai20}  
\usepackage{times}  
\usepackage{helvet} 
\usepackage{courier}  
\usepackage[hyphens]{url}  
\usepackage{graphicx} 
\usepackage{graphicx}
\usepackage{array}
\usepackage{amsfonts}
\usepackage{amsmath}
\usepackage{adjustbox} 
\title{Using Chinese Glyphs for Named Entity Recognition} 
\author  {Arijit Sehanobish\thanks{equal contribution} \thanks{Work done during SCALE program at JHU}
\\ 
Yale University\\ arijit.sehanobish@yale.edu
\And Chan Hee Song\footnotemark[1] \footnotemark[2] \\ University of Notre Dame \\ csong1@nd.edu 
}

\begin{document}

\maketitle

\begin{abstract}
    Most Named Entity Recognition (NER) systems use additional features like part-of-speech (POS) tags, shallow parsing, gazetteers, etc. Adding these external features to NER systems have been shown to have a positive impact. However, creating gazetteers or taggers can take a lot of time and may require extensive data cleaning. In this work instead of using these traditional features we use lexicographic features of Chinese characters. Chinese characters are composed of graphical components called radicals and these components often have some semantic indicators. We propose CNN based models that incorporate this semantic information and use them for NER. Our models show an improvement over the baseline BERT-BiLSTM-CRF model.
    We present one of the first studies on Chinese OntoNotes v$5.0$ and show an improvement of $+.64$ F1 score over the baseline. We present a state-of-the-art (SOTA) F1 score of $71.81$ on the Weibo dataset, show a competitive improvement of $+0.72$ over baseline on the ResumeNER dataset, and a SOTA F1 score of $96.49$ on the MSRA dataset.
\end{abstract}

\section{Introduction}

    \indent Augmenting named entity recognition (NER) systems with additional features like gazetteers, bag of words or character level information has been commonplace like Sang and Meulder (2003), Collobert et al.~(\citeyear{Collobert}) and Chiu and Nichols~(\citeyear{chiu2016}). Over the years these added features have shown to improve the NER systems. We follow this approach to the Chinese NER task. Chinese is a logographic language and many Chinese characters have evolved from pictures, thus we incorporate the semantic information from the pictures of the Chinese characters which can be used as an added feature for Chinese NER systems. Several authors \citeauthor{Shi}~(\citeyear{Shi}), \citeauthor{Li}~(\citeyear{Li}), \citeauthor{Sun}~(\citeyear{Sun}) and Meng et al.~(\citeyear{Glyce}) managed to use the radical representations successfully in a wide range of natural language understanding (NLU) tasks. However using these images for NLU have sometimes proved to be a challenge. For example \citeauthor{Liu}~(\citeyear{Liu}), and Zhang and LeCun~(\citeyear{Zhang}) work shows inconsistent results and some even yield negative results (Dai and Cai~\citeyear{Dai}). Furthermore, there has not been a lot of work in using glyphs particularly for NER. Only recently,  \citeauthor{Glyce}~(\citeyear{Glyce}) presented a complex Glyph reinforced model that concatenates glyph embeddings with BERT (Devlin et al.~\citeyear{Devlin}) embeddings. In this paper, we present a new kind of glyph-augmented architecture that is easier to implement and train, more robust and requires less data. We show that our models have a significant improvement over our baseline on two datasets, Chinese OntoNotes v$5.0$ (\citeauthor{pradhan2017ontonotes}~\citeyear{pradhan2017ontonotes}) and Weibo~(\citeauthor{Weibo}~\citeyear{Weibo}).\\
    \indent We approach the problem of incorporating Chinese glyphs as an image classification problem and present two CNNs which we call ``strided" and ``GLYNN" inspired from computer vision. We treat this encoding problem purely in terms of computer vision, i.e. to extract ``meaningful" features from the image, instead of a specialized CNN that encapsulates the subtle radicals. Both CNNs are used to encode the glyphs and these encoded images are then used as an added feature for our NER system. We also present an autoencoder architecture to pretrain GLYNN and compare the results. Due to treating the problem as an image classification problem, our models are easier to train and implement.  Since the OntoNotes v$5.0$ and Weibo datasets have a significant number of non-Chinese characters in them, we also show our model is robust by conducting a robustness test of our systems by throwing in the pictures of the non-Chinese characters as well pictures of English alphabets and show that it still beats the baseline model. English alphabets unlike Chinese have no semantic information so we can think of these added pictures as ``noise". Furthermore, unlike \citeauthor{Glyce}~(\citeyear{Glyce}) who used extensive dataset of Chinese glyphs gathered from different sources, we use only about $4500$ grayscale $64$ x $64$ Chinese characters in Hei Ti font. Hei Ti font, similar to sans-serif, is widely used and is easy to gather glyph information. This enables us to use a lot less data to train our model. In this paper, we found that all our models are an improvement over the baseline and achieve state of the art F1 score on the Weibo dataset. Our code can be found in https://github.com/arijitthegame/BERT-CNN-models. The main strength of our model is as follows: 
    \begin{itemize}
    \item Easy to implement and train
    \item Robust to non-Chinese languages in the dataset
    \item Requires less amount of glyph data to train
    \end{itemize}

\section{Architecture of our GLYPH models}

\begin{figure}[!h]
\centering
\includegraphics[width=\columnwidth]{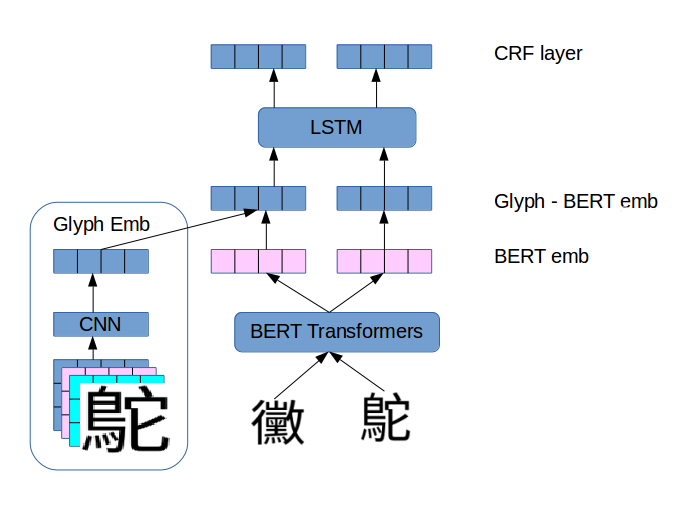}
\caption{Architecture of our Glyph Models}
\label{figure: 1}
\end{figure}

BERT is the state of the art language model introduced by \citeauthor{Devlin}~(\citeyear{Devlin}). BERT is a transformer based model which is trained on masked word prediction and next sentence prediction tasks and is trained on Wikipedia data and a large book corpus. We use the Chinese BERT base which is released by Google and is publicly available on Google Github. We then combine BERT with a popular used architecture in NER, BiLSTM-CRF as in Huang, Xu and Yu~(\citeyear{huang2015bidirectional}), Ma and Hovy~(\citeyear{ma-hovy}), Chiu and Nichols~(\citeyear{chiu2016}) and Zhang and Yang~(\citeyear{latticeLSTM}).   \\
Our model (figure~\ref{figure: 1}) consists of the following parts: pretrained BERT embeddings and the (pretrained) CNN embeddings. We concatenate the last four layers of BERT and the CNN vectors which are our new ``character" embeddings. We then feed these character embeddings to the BiLSTM layer which are finally decoded via a CRF layer. The CNN-LSTM-CRF is then being trained end-to-end while we keep BERT frozen. Thus we try to take advantage of BERT's large pre-scale training and the information from Chinese glyphs encoded by our CNN's. 

The selection of the CNN's are primarily motivated by problems in computer vision. The convolution layers are the key in extracting meaningful features and one needs at least two to extract meaningful features.  Batch normalization~\cite{batch} and layer normalizations~(Ba, Kiros and Hinton~\citeyear{Ba}) are generally used to accelerate training. To vindicate our choice of these CNN's we applied them to ``Fashion MNIST" dataset and we got around $93\%$ accuracy.

\subsection{Obtaining Glyph information} Before giving a detailed description of the CNNs used to extract glyph features we want to explain how the model is trained. All the datasets used in this paper are character tokenized and in IOB or BIOES format. We construct a dictionary of $4200$ Chinese characters where the key is the UTF-$8$ codepoint and the value is the numpy array of pixels. A character in each line is one of the following types:
\begin{itemize}
\item Chinese character and one of the $4200$ characters we used: We look up it’s image via our dictionary using it’s UTF-$8$ codepoint and pass the image through our CNN (strided or GLYNN, depending on the model).
\item Out of vocabulary Chinese characters (there are about a hundred of them across various datasets): Since we don’t have that character’s picture we use a black image, i.e. $64x64$ pixel of all $1$'s and then pass the black image through the CNN.
\item Non Chinese character: we use a white image, i.e. $64x64$ pixel of all $0$’s and pass it through our CNN.
\end{itemize}
In each of the above cases, we then get the BERT output of the character and concatenate the BERT output with the output of the CNN. These concatenated vectors are our character representations. 

\subsection{Strided CNN}
\begin{figure}[!h]
\centering
\includegraphics[width=\columnwidth]{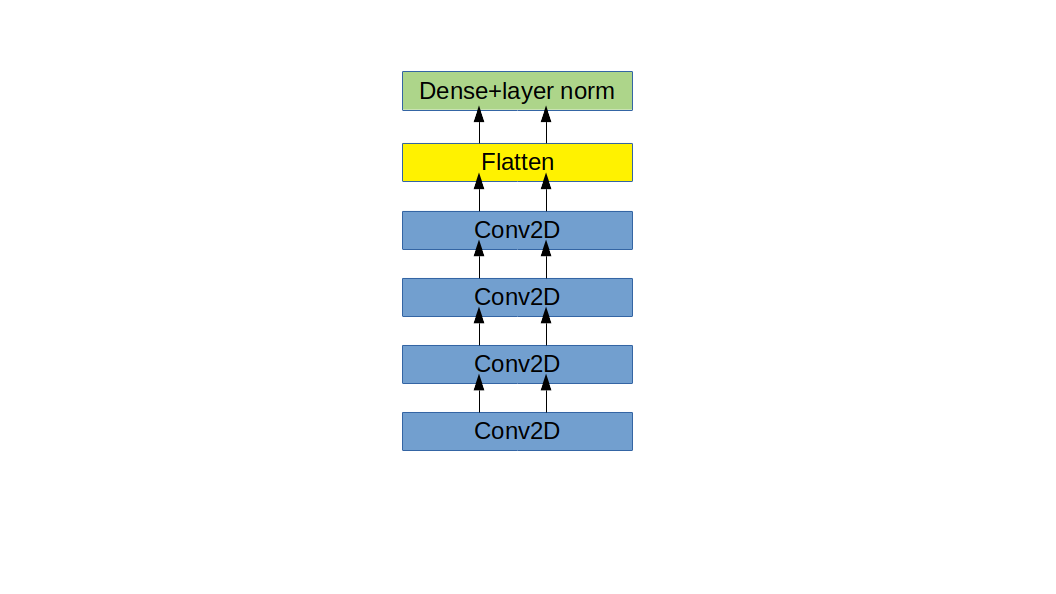}
\caption{Strided CNN encoder}
\label{figure: 2}
\end{figure}

This CNN is inspired from the work of Su and Lee~(\citeyear{Su}) who used $5$ convolution layers. They used their CNN for Chinese word analogy problems. We did an ablation study and adapted a smaller CNN for our model. Figure~\ref{figure: 2} is the picture of our strided CNN. This consists of $4$ $2$D convolution layers with strides $2$, filter size of $64$, kernel size of $3$ and activation leaky ReLU. Then we have a Flatten layer and a Dense layer of output dimension $64$. Furthermore we normalize the final output by using layer normalization as introduced by \citeauthor{Ba}~(\citeyear{Ba}).

\subsection{GLYNN CNN}
\begin{figure}[!h]
\centering
\includegraphics[width=\columnwidth]{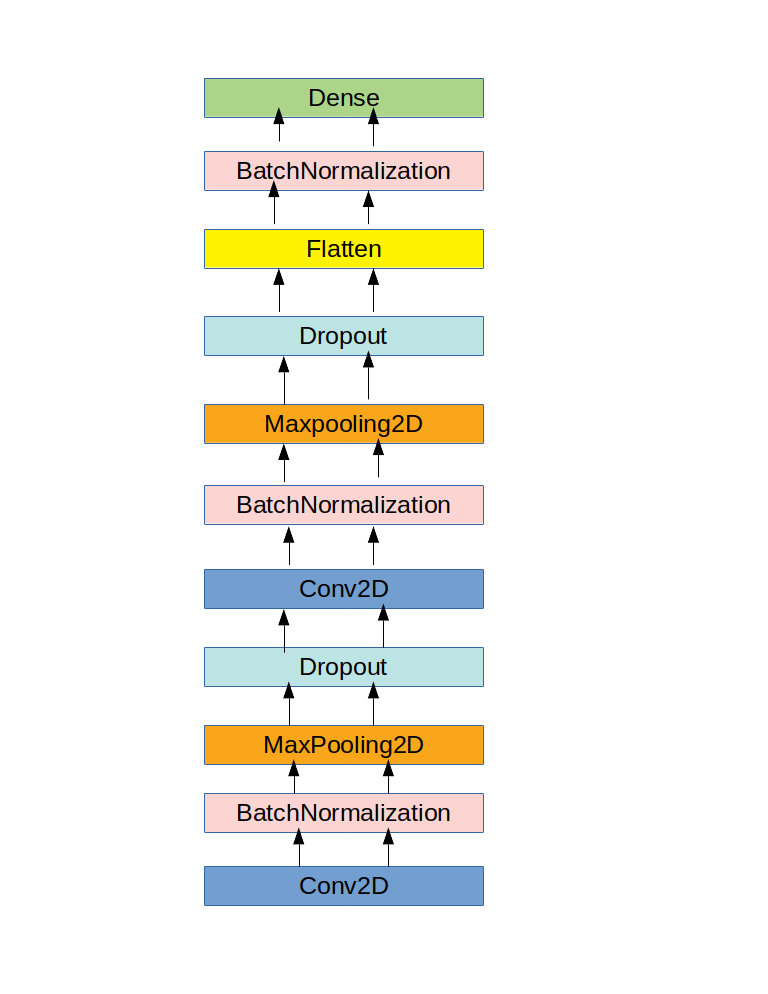}
\caption{GLYNN CNN Encoder}
\label{figure: 3}
\end{figure}

In this subsection we describe another CNN which we call ``Glynn" to encode the glyphs. The idea for this CNN comes purely from image classification tasks. The convolution layers are the most important component of CNN and one needs at least $2$ layers to capture details of an image. Batch normalization (Ioffe and Szegedy 2015) is used to speed up training and the dropout layers are used to prevent overfitting and the maxpooling layers are used to reduce the computational complexity of the network. Figure~\ref{figure: 3} is the picture of our CNN.

\begin{figure}[!hb]
\centering
\includegraphics[width=\columnwidth]{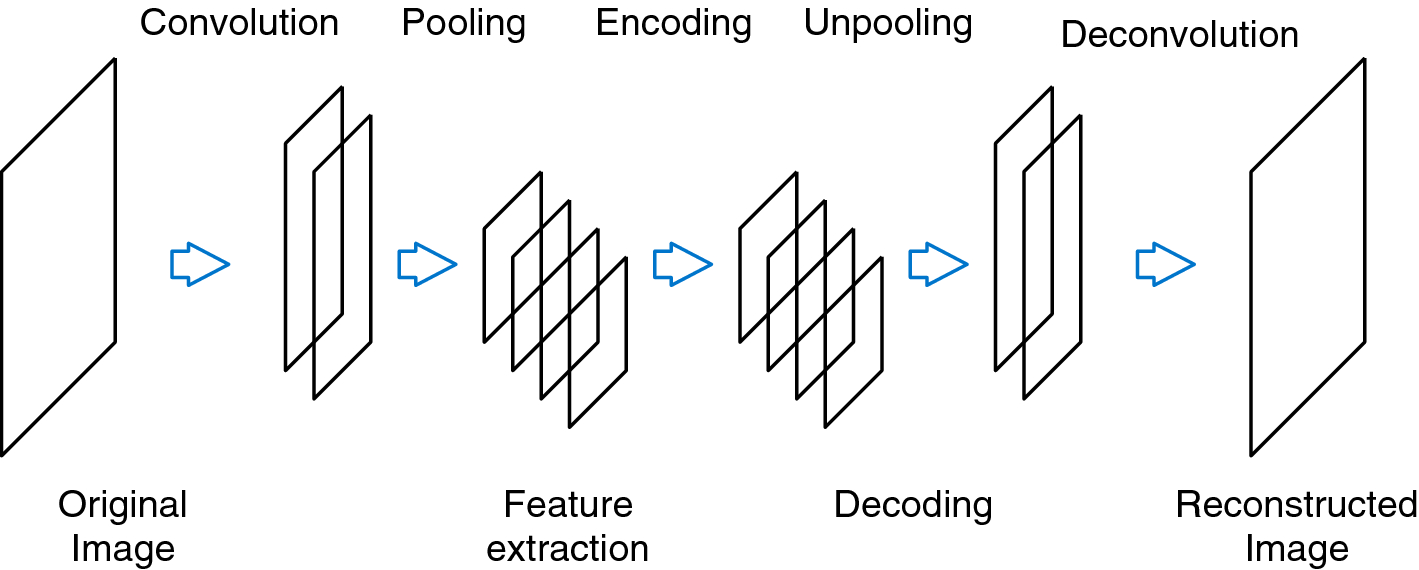}
\caption{Autoencoder}
\label{figure: 4}
\end{figure}

 We use filter size of $32$, kernel size of $3$ and padding=`same' in both the convolution layers, while we use sigmoid activation and strides=$(2,2)$ for our first convolution layer and ReLU activation and strides=$(1,1)$ for the second convolution layer. For both the maxpooling layers we use a pool size of $2$. And finally we use two dropout layers. The output dimension of the CNN is $256$. We pretrain the CNN using an autoencoder shown in figure~\ref{figure: 4}.

\subsection{Autoencoder}

We also employ autoencoder (\citeauthor{AE}~\citeyear{AE}) to pretrain the CNN. The main purpose of the autoencoder is dimensionality reduction, i.e. the autoencoder encourages the CNN to extract high level features without employing multiple convolution layers like the strided CNN. A deep autoencoder architecture is a multi-layer neural network that tries to reconstruct it's input. The architecture therefore consists of a sequence of $N$ encoder layers followed by a sequence of $N$ decoder layers. Let $\Theta := \{\Theta^i, 1 \leq i \leq N \}$ be the parameters of each layer, where each $\Theta^{i}$ can be further written as $\Theta^{i} = \{ W_{e}^{i}, b_{e}^{i}, W_{d}^{i}, b_{d}^{i} \}$, where $W_{e}^{i}, b_{e}^{i}$ are the weights and the biases of the $i$-encoder layer and $W_{d}^{i}, b_{d}^{i}$ are the weights and the biases of the $i$-decoder layer. If $x \in \mathbb{R}^d$ be an input, then the encoder-decoder architecture can be formulated by the following series of equations. 
\begin{align}
\label{eqn: eqlabel}
\begin{split}
 h^{0} & := x  ,
 \\
 h^{i} & = f_{e}^{i}(h^{i-1}) := s_e^{i} (W_{e}^{i}h^{i-1} + b_{e}^{i}),
 \\
 h_{r}^{n} & := h^n ,
 \\
  h_{r}^{i} & = f_{d}^{i}(h^{i+1}) := s_d^{i} (W_{d}^{i}h_{r}^{i+1} + b_{e}^{i})
 \end{split}
\end{align}
where $f_{e}^{i}$ and $f_{d}^{i}$ are the encoding and decoding functions and $s_{e}^{i}$ (resp. $s_{d}^{i}$) are elementwise non-linear functions and $h_{r}^{1} \in \mathbb{R}^{d}$. The training objective is to find a set of parameters $\Theta^{opt} := \{ \mathbb{W}_e , \mathfrak{b}_e , \mathbb{W}_d, \mathfrak{b}_d \}$ which will minimize the reconstruction error over all points in the dataset $\mathfrak{D}$, i.e. 
\begin{equation}
    \mathfrak{L} : = \frac{1}{|\mathfrak{D}|} \sum_{x \in \mathfrak{D}} ||x - h_{r}^{1}(x)||_{2}^{2}
\end{equation}
and $|| \cdot||$ is the usual $L^{2}$-norm in $\mathbb{R}^{d}$. \newline 
We train the autoencoder for $200$ epochs and we use RMSprop as our optimizer. \\

In the GLYNN CNN we take the encoder part of the autoencoder,

\section{Related Works}

Our work follows the mainstream approach to named entity recognition. Mainstream neural approach predates to 2003 when \citeauthor{Hammerton2003}~(\citeyear{Hammerton2003}) used Long Short-Term Memory for NER, achieving just above average for English F1 scores and improvement for German NER. \citeauthor{Hochreiter:1997:LSM:1246443.1246450}~(\citeyear{Hochreiter:1997:LSM:1246443.1246450}) presented Long Short-Term Memory (LSTM), and it was expanded by Gers, Schmidhuber and Cummings~(\citeyear{Gers2000}), and reached its current form by Graves and Schmidhuber~(\citeyear{Graves05}). LSTM is increasing in its use with NER problems over the past $2$ decades. Recent works in NER follows this approach, mainly using BiLSTM-CRF architecture. Bi-LSTM-CRF architecture was first proposed by Huang, Xu and Yu~(\citeyear{huang2015bidirectional}), and has been widely studied and augmented. Chiu and Nichols~(\citeyear{chiu2016}) and Ma and Hovy~(\citeyear{ma-hovy}) augmented LSTM-CRF architecture with character-level convolutional neural network to add an additional features to the architecture. Instead of applying convolutional neural network to the text, we apply it to the glyphs to augment our Bi-LSTM-CRF. \\
\indent Recently, transfer learning architectures has shown significant improvement in various natural language processing tasks such as question answering, natural language understanding, machine translation and natural language inference. \citeauthor{Devlin}~(\citeyear{Devlin}) uses stacked bi-directional transformer layers called BERT that is trained on masked word prediction and next sentence prediction tasks. BERT is trained on over $3,300$M words mostly gathered from Wikipedia. By employing a task-specific final output layer, BERT can be tuned to many different natural language processing tasks. In this work, we apply BERT to NER and use BiLSTM-CRF as the output layer of BERT-CNN. Our approach presents an architecture that doesn't require a dataset-specific architecture and feature engineering.
\newline
\indent In the recent years there has been a lot of work to use Chinese glyphs as an added feature for various language understanding tasks. Our work is similar to Meng et al~(\citeyear{Glyce}) but differs in many aspects. Unlike Meng et al, who uses ensemble of glyph images from different time periods and writers, which is often hard to collect. Instead we use only about $4500$ grayscale $64$ x $64$ Chinese characters in Hei Ti font. Hei Ti font, which is similar to sans-serif, is widely used and easy to collect glyph data. These are all the Chinese characters found in Chinese BERT vocabulary. Even though there are over $20,000$ CJK characters, we only have about a hundred of out-of-vocabulary characters in OntoNotes v$5.0$ and Weibo. So that allowed us to use considerable less data than Meng et al.~(\citeyear{Glyce}). Another major difference between our approach and all the aforementioned authors in the introduction is that our CNN's are agnostic to the subtleties of Chinese characters and we treat this encoding problem with computer vision ideas and extract ``meaningful" features from the image, instead of having a specialized CNN that encapsulates the subtle radicals. Both Su and Lee~(\citeyear{Su}) and Meng et al~(\citeyear{Glyce}) use autoencoders to pretrain their CNN. Su and Lee~(\citeyear{Su}) pretrain the CNN by freezing some layers while Meng et al~(\citeyear{Glyce}) pretrain the CNN with the objective of recovering an ``image id" while we follow the approach of jointly training all layers of the CNN \cite{AE} with the global objective of reconstructing the image. We also employ pretraining to GLYNN and shows that it has a performance gain compared to not pretraining. Another difference between our architecture and the GLYCE model (Meng et al.~\citeyear{Glyce}) is that we use a BiLSTM instead of a transformer to encode the BERT + Glyph Embeddings. Overall, our model is a very robust and easy to train model that uses very little data for augmenting the glyph features and successfully marry techniques from computer vision and state of the art language models.

\section{Experimental results}

\begin{table}[h!]
    \centering
    \begin{tabular}{|c|c|c|c|c|}
    \hline
    & & \textbf{Train} & \textbf{Dev} & \textbf{Test} \\
    \hline
        OntoNotes v$5.0$ & tok & $1.2$M & $178$K & $149$K \\
        & ent & $65$K & $9401$ & $7785$ \\
        & sent & $37$K & $6217$ & $4293$ \\
    \hline
        Weibo & tok & $1855$ & $379$ & $405$ \\
        & ent & $957$ & $153$ & $211$ \\
        & sent & $1350$ & $270$ & $270$ \\
    \hline
    ResumeNER & tok & $124.1$k & $13.9$k  & $15.1$k \\
    & ent & $13.4$k & $1497$ & $1630$ \\
    & sent & $3.8$k & $0.46$k & $0.48$k \\
    \hline
    MSRA & tok & $2169.9$k & - & $172.6$k \\
    & ent & $75$k & - & $6.1$k \\
    & sent &  $46.4$k & -& $4.4$k \\
    \hline
    \end{tabular}
    \caption{Statistics of the OntoNotes v$5.0$, Weibo, MSRA and ResumeNER dataset, \textit{tok} stands for number of tokens in each split, \textit{ent} stands for number of annotated entities in each split and \textit{sent} stands for number of sentences in each split}
    \label{table: 1}
\end{table}

\subsection{Datasets used} 
We used Chinese OntoNotes v$5.0$ dataset compiled for CoNLL-2013 shared task~\cite{pradhan2017ontonotes} and follow the standard train/dev/test split as presented in Pradhan et al.~(\citeyear{pradhan2013ontonotes}). OntoNotes v$5.0$ is composed of $18$ different tag sets, ART, DAT, EVT, FAC, GPE, LAW, LNG, LOC, MON, NRP, NUM, ORD, ORG, PCT, PER, PRD, QTY and TIM. We chose OntoNotes v$5.0$ due to it being comprehensive of previous OntoNotes releases. OntoNotes v$5.0$ contains an extra genre, telephone communications, which makes the dataset more representative of the real world. We also use Weibo dataset~\cite{Weibo} and ResumeNER dataset~\cite{latticeLSTM} which  has $4$ entity types: PER, ORG, GPE and LOC and $8$ entity types: CONT, EDU, LOC, NAME, ORG, PRO, RACE and TITLE respectively. We use both named entity mention and nominal mentions for the Weibo dataset as well as Weibo with just the named entity mention which we refer to by Weibo NAM. MSRA dataset has $3$ entity types PER, LOC and ORG. Since MSRA do not have a dev set, we take 10\% of the test set as our dev set. The statistics of datasets are shown in Table~\ref{table: 1}. We use the train set to train the model, the dev set for validation and the test set for testing.
All the results are results from the test set. 

\subsection{NER results on OntoNotes v$5.0$, Weibo and ResumeNER}
 To make a proper comparison, we run our vanilla BERT-LSTM-CRF models $10$ times on OntoNotes v$5.0$ and $20$ times on Weibo NAM and $40$ times on Weibo to establish a baseline. For better understanding of our results, in Table~\ref{table: 2} we give a complete list of all hyperparameters used in running these experiments. 
 \newline
 \begin{table}[h!]
\begin{center}
\adjustbox{max width=\columnwidth}{
\begin{tabular}{|p{10em}|p{8em}|}
\hline
\multicolumn{2}{|c|}{Hyperparameters}\\
\hline
Number of BiLSTM layers & $1$ \\
\hline
Hidden\_size\_LSTM & $256$ \\
\hline
dropout\_LSTM & $.5$ \\
\hline
optimizer & adafactor \\
\hline
clip\_grad\_norm & $1$ \\
\hline
learning rate scheduler & cosine\_decay \& first decay steps = $1000$ \\
\hline
BERT layers used & $-4,-3$, $-2,-1$ \\
\hline
weight\_decay & $.005$ \\
\hline
\raggedright Default dropout in GLYNN CNN & $.3$, $.5$ resp. \\
\hline
training epochs & $30$ \\
\hline
mini batch size & $8$ \\
\hline
\end{tabular}
}
\caption{Default hyperparameters used in the experiments}
\label{table: 2}
\end{center}
\end{table}

 We ran $10$ trials for $30$ epochs for each of Glynn CNN (default and higher dropout) and strided CNN on Chinese OntoNotes v$5.0$. We also compare the statistical significance of our results over the baseline by performing a $2$-sample t-test. We call a result statistically significant if the $p$-value is less than $.05$. We report our scores in Table~\ref{table: 3}.
\newline
\begin{table}[h!]
\centering
\adjustbox{max width = \columnwidth}{
\begin{tabular}{|c|c|c|c|c|c|}
\hline
Models & Avg & Std dev & Max & p-value
 \tabularnewline
\hline
BERT-BiLSTM-CRF (baseline) & $78.80$ & $.33$ & $79.09$ & N/A
\tabularnewline
\hline
GLYNN & $79.24$ & $.16$ & $79.47$ & $.004$ \\
GLYNN +dropout .5 &$79.22$ & $.13$ & $79.35$ & $.005$\\
strided & $79.59$ & $.12$ & $\mathbf{79.73}$ & $<.001$
\tabularnewline
\hline
Che et al. ($2013$) & N/A & N/A & $69.82$ & N/A \\
\hline
Pappu et al. ($2017$) & N/A & N/A & $67.2$ & N/A \\
\hline
Xu et al. ($2017$) & N/A & N/A & $71.83$ & N/A \\
\hline
Nosirova et al. ($2019$) & N/A & N/A & $72.12$ & N/A \\
\hline
\end{tabular}
}
\caption{Results on OntoNotes v$5.0$}
\label{table: 3}
\end{table}

Using the F1 scores obtained by the Glynn CNN (dropout $.5$) and the strided CNN, we perform the $2$- sample t-test and obtain a $p$-value of $<.001$.  Thus we see that strided CNN is a significant improvement over both the BERT baseline and the GLYNN. But interestingly both the GLYNN models outperform strided on the dev sets. Finally we note that it is not a huge surprise that the gains are low on OntoNotes v$5.0$ as the dataset is big and we used a small amount of data to augment the system.
\newline
\indent Since the Weibo dataset is smaller and is noisier than OntoNotes v$5.0$, we ran vanilla BERT-BiLSTM-CRF $20$ times and $40$ times respectively on Weibo NAM and the Weibo dataset to establish a baseline. We ran $20$ trials for $30$ epochs for each of Glynn CNN (default and higher dropout) and strided CNN on Weibo. We also calculate the $p$-value between our CNN's and the baseline BERT to see if our results are statistically significant. Table~\ref{table: 4} gives us a summary of our results on Weibo and we compare our results with He and Sun~(\citeyear{oldweibo}). Table~\ref{table: 4} shows that Weibo experiment results from both CNNs are statistically significant than the vanilla BERT. However $p$- value between strided CNN and GLYNN is $.71$, which shows that their performance on average is statistically the same. 
\newline
\begin{table}[h!]
\centering
\adjustbox{max width=\columnwidth}{

\begin{tabular}{|c|c|c|c|c|}
\hline
Models & Avg & Std dev & Max & p-value 
 \tabularnewline
\hline
BERT-BiLSTM-CRF (baseline) & $69.9$ & $1.4$ & $71.8$ & N/A
\tabularnewline
\hline
GLYNN & $71.44$ & $1.48$ & $\mathbf{73.9}$ & $.003$
\tabularnewline
GLYNN + dropout .5 & $71.34$ & $1.2$ & $73.36$ & $.003$ \\
strided & $71.34$ & $1.48$ & $73.35$ & $.005$ 
\tabularnewline
\hline
\citeauthor{oldweibo}~(\citeyear{oldweibo}) & $54.50$ & N/A & N/A & N/A
\tabularnewline
\hline
\citeauthor{zhu-wang-2019-ner}~(\citeyear{zhu-wang-2019-ner}) & $55.38$ & N/A & N/A & N/A \\
\hline
\end{tabular}}
\caption{Results on Weibo NAM}
\label{table: 4}
\end{table}

Table~\ref{table: Weibo} shows our results on the full Weibo dataset. Even though strided CNN did not show any improvement, we made significant gains with the GLYNN CNN over the baseline and set a new SOTA F1 score. Even though strided CNN shows improvement over the baseline on other datasets and the dev sets, we do not understand why it underperforms in this case. We would like to investigate this problem further. 

\begin{table}[h!]
\centering
\adjustbox{max width=\columnwidth}{
\begin{tabular}{|c|c|c|c|c|}
\hline
Models & Avg & Std dev & Max & p-value 
 \tabularnewline
\hline
BERT-BiLSTM-CRF (baseline) & $68.68$ & $1.07$ & $70.79$ & N/A
\tabularnewline
\hline
GLYNN & $69.2$ & $1.11$ & $\mathbf{71.81}$ & $.03$
\tabularnewline
GLYNN + dropout .5 & $69.01$ & $1.11$ & $71.58$ & $.18$ \\
strided & $68.67$ & $.97$ & $70.7$ & $.97$ 
\tabularnewline
\hline
\citeauthor{Glyce}~(\citeyear{Glyce}) & N/A & N/A & $67.6$ & N/A
\tabularnewline
\hline
\citeauthor{latticeLSTM}~(\citeyear{latticeLSTM}) & N/A & N/A & $58.79$ & N/A \\
\hline
\end{tabular}}
\caption{Results on Full Weibo}
\label{table: Weibo}
\end{table}

Table~\ref{table: ResumeNER} shows our results on the ResumeNER dataset. We ran our experiments $15$ times and we hope a grid search will make our results closer to SOTA scores reported by \citeauthor{Glyce}~(\citeyear{Glyce}). We would also like to point out that we trained our models for $20$ epochs which seemed to give the best results. 
\begin{table}[h!]
\centering
\adjustbox{max width=\columnwidth}{
\begin{tabular}{|c|c|c|c|c|}
\hline
Models & Avg & Std dev & Max & p-value 
 \tabularnewline
\hline
BERT-BiLSTM-CRF (baseline) & $94.92$ & $.51$ & $95.72$ & N/A
\tabularnewline
\hline
GLYNN & $95.61$ & $.62$ & $96.49$ & $.02$
\tabularnewline
strided & $95.66$ & $.47$ & $96.42$ & $.01$ 
\tabularnewline
\hline
\citeauthor{Glyce}~(\citeyear{Glyce}) & N/A & N/A & $\mathbf{96.54}$ & N/A
\tabularnewline
\hline
\citeauthor{latticeLSTM}~(\citeyear{latticeLSTM}) & N/A & N/A & $94.46$ & N/A \\
\hline
\end{tabular}}
\caption{Results on ResumeNER}
\label{table: ResumeNER}
\end{table}

Table~\ref{table: MSRA} shows our results on MSRA. We ran $10$ trials for each experiment for $30$ epochs. 

\begin{table}[h!]
\centering
\adjustbox{max width=\columnwidth}{
\begin{tabular}{|c|c|c|c|c|}
\hline
Models & Avg & Std dev & Max & p-value 
 \tabularnewline
\hline
BERT-BiLSTM-CRF (baseline) & $95.05$ & $.41$ & $95.3$ & N/A
\tabularnewline
\hline
GLYNN & $95.07$ & $.48$ & $\mathbf{96.49}$ & $.8$
\tabularnewline
strided & $95.21$ & $.42$ & $95.63$ & $.67$ 
\tabularnewline
\hline
\citeauthor{Glyce}~(\citeyear{Glyce}) & N/A & N/A & $95.54$ & N/A
\tabularnewline
\hline
\citeauthor{latticeLSTM}~(\citeyear{latticeLSTM}) & N/A & N/A & $93.18$ & N/A \\
\hline
\end{tabular}}
\caption{Results on MSRA}
\label{table: MSRA}
\end{table}

\subsection{Robustness test}
$\frac{1}{9}$ of Chinese OntoNotes v$5.0$ and $\frac{1}{6}$ of Weibo characters are not Chinese. So a natural question would be: what if we add the pictures of the other non-Chinese characters as well. To test the performance of the system, we added in an additional $3000$ pictures. These are the pictures of all non CJK characters found in the Chinese BERT vocabulary. Of course, a picture of an English character has no semantic meaning so adding these characters can be thought as a robustness test of our NER systems. However we do not change the number of Chinese characters used. Now the main differences are as follows: 
\begin{itemize}
    \item We construct a larger dictionary ($8000$ key-value pairs).
\item If the character’s codepoint is a key in our dictionary, we look up it’s image and pass through our CNN as before.
\item If the codepoint is a non Chinese character and not a key, we then use the white image as before.
\item If the codepoint is a Chinese character and not a key, then we use the black image as before.

\end{itemize}

Over $200$ trials with various hyperparameters, we found that adding the pictures of non-Chinese characters drop the F1 scores but the models still beat the baseline model. In the tables~\ref{table : 5},~\ref{table : 6} we show a snippet of our results. The results in the Table~\ref{table : 5} are compiled over $20$ trials running for $30$ epochs with the default hyperparameters as in Table~\ref{table: 2}.

\begin{table}[h!]
\centering
\begin{tabular}{|c|c|c|c|}
\hline
Models & Avg & Std dev & Max \\ 
\hline
GLYNN + dropout $.5$ & $70.83$ & $1.18$ & $73.61$ \\ 
\hline
GLYNN & $70.7$ & $1.55$ & $73.25$ \\
\hline
strided & $70.7$ & $1.6$ & $73.81$\\
\hline
\end{tabular}
\caption{Results on Weibo NAM with added pictures}
\label{table : 5}
\end{table}

However the difference in performance between GLYNN (with $.5$ dropout) and strided is statistically insignificant as the p-value is $.768$. \\

In the Table~\ref{table : 6} below we show some our results on OntoNotes v$5.0$ compiled over $10$ trials running for $30$ epochs with the default hyperparameters as in Table~\ref{table: 2}. 

\begin{table}[h!]
\centering
\begin{tabular}{|c|c|c|c|}
\hline
 & Avg & Std dev & Max \\ 
\hline
GLYNN + dropout $.5$ & $79.31$ & $.19$ & $79.59$  \\ 
\hline
GLYNN & $79.27$ & $.04$ & $79.33$ \\
\hline
strided & $78.95$ & $.26$ & $79.21$\\
\hline
\end{tabular}
\caption{Results on OntoNotes v$5.0$ with added pictures}
\label{table : 6}
\end{table}

We believe the reason for the diminished performance is due to the fact that the pictures of various extra characters do not have any semantic meaning unlike the Chinese characters. So these pictures give a noisy signal to our NER system which in turn affects the performance. 

\subsection{Hyperparameter tuning}
In this subsection we will discuss other hyperparameters we try out during our experiments. 
\\
\\
\textbf{Effects of learning rates and optimizers :} We used Adam optimizer on OntoNotes v$5.0$ for $5$ trials with the learning rates $.001$, $.0005$ and $.0001$ with early stopping if the loss did not decrease over 5 epochs. We obtained an average F1 scores of $77.8$, $76.64$ and $77.72$ respectively. But in general, we found Adam performs poorly compared to Adafactor. \\
Early stopping with all the above learning rates with both Adam and Adafactor on Weibo produced erratic results with extremely high standard deviations.\\
\\
\textbf{Effects of dropout :}
We also used dropouts of $.5$ on each the dropout layers of the Glynn CNN and ran multiple trials. Table~\ref{table: 3} summarizes our results on OntoNotes v$5.0$ after $10$ trials and we also compute the $p$-value to test the difference in the average performance of GLYNN.
\newline
Changing the dropouts to $.5$ did not have any statistically significant improvement over the default hyperparameters on OntoNotes v$5.0$ and ResumeNER. \\
We also ran $20$ trials on Weibo with this new dropout. The results are in Table~\ref{table: 4}.
We did $2$ sample t-test between GLYNN with $.5$ dropouts and the strided CNN (resp. GLYNN and GLYNN with $.5$ dropouts) and we found the $p$-value to be $.84$ (resp $.83)$. So the two CNN (with or without higher dropout) behave pretty much the same. However on the dev set, we found that the higher dropouts tend to do better for Weibo.\\
\\
\textbf{Effects on changing the training epochs :}
We also ran Weibo NAM on $20$ and $40$ epochs but running on $30$ epochs gives us the best results. Running Weibo NAM for $20$ epochs shows a drop in the F1 scores of the LOC tags and thus results in a very poor performance. Losses tend to increase after $30$ epochs and so the models start doing worse at $40$ epochs. $30$ is also an optimum choice for the full Weibo dataset and OntoNotes. We found that running ResumeNER for $20$ epochs yield better results.

\begin{figure}[!h]
\centering
\includegraphics[width=\columnwidth, height=6cm]{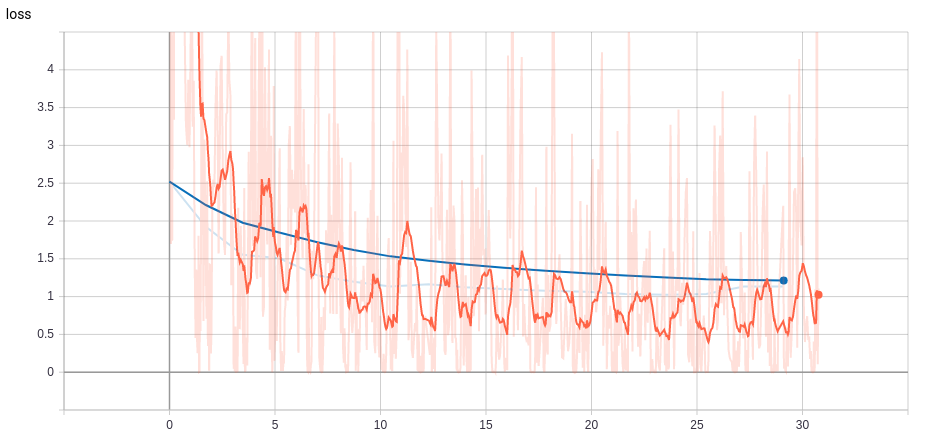}
\caption{GLYNN loss on OntoNotes on various epochs with optimizer Adafactor}
\label{figure: 5}
\end{figure}

\begin{figure}[!h]
\centering
\includegraphics[width=\columnwidth]{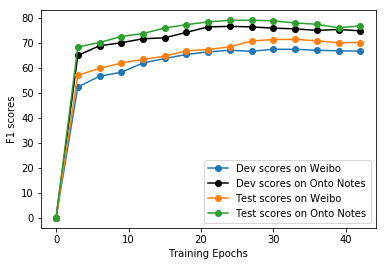}
\caption{Training GLYNN on various epochs on Weibo NAM and OntoNotes v$5.0$}
\label{figure: 6}
\end{figure}

But we found that lowering the epoch number hurt the performance significantly. We also trained all our models with an early stopping if the loss did not decrease over $3$ and $5$ epochs as well. That resulted in our models running on OntoNotes to stop around $12$ and $18$ epochs respectively. Figure~\ref{figure: 5} shows the relationship between training epochs and loss where the red is the training loss and the blue is the validation loss. Figure~\ref{figure: 6} shows the relation between training epochs of GLYNN and the test and dev F1 scores on Weibo NAM and OntoNotes.

\noindent\textbf{How important is the autoencoder :} We ran multiple tests ($20$ for Weibo NAM, $10$ for OntoNotes v$5.0$) with varying learning rates and training epochs. We found the autoencoder improved performance slightly and reduced variance on the dev and test sets. For example, GLYNN with default hyperparameters without the autoencoder got an average test score of $70.98 (\pm 1.47)$ on Weibo NAM.   

\section{Conclusion and Future Work} 
Using two very different CNNs with and without an autoencoder, we have shown gains over the baseline system on the three most commonly used datasets and achieve state of the art F1 score on the Weibo dataset. The novelty of our approach lies in $3$ salient features: a) very little data to augment our system, b) it's easier to train and implement and c) robust. Generating glyph images from text is also not a time consuming process and our models require less glyph data to train. Thus our hope is that using glyphs as an added feature will become a more commonplace occurrence for Chinese NER as it is an easy and quick method to improve the NER systems. We are excited by the future of glyphs in NLP and we would like use glyphs for other NLP tasks. 

\section{Acknowledgements} 
We would like to thank Johns Hopkins and the SCALE workshop for their hospitality and for facilitating an excellent atmosphere for conducting research. We would like to thank Nicholas Andrews for being an excellent mentor and for his continuous support and guidance. We also would like to thank Dawn Lawrie for her mentorship in this project. Finally, we thank Jim Mayfield for preparing the datasets, Derek Zhang for running various experiments, David Etter for the images of the characters, Oyesh Singh and David Mueller for answering our questions. 
\nocite{*}
\bibliography{Bibliography.bib}

\begin{thebibliography}{}

\bibitem[\protect\citeauthoryear{Ba, Kiros, and Hinton}{2016}]{Ba}
Ba, J.~L.; Kiros, J.~R.; and Hinton, G.~E.
\newblock 2016.
\newblock Layer normalization.
\newblock \url{https://arxiv.org/abs/1607.06450}.
\newblock preprint.

\bibitem[\protect\citeauthoryear{Che \bgroup et al\mbox.\egroup }{2013}]{che}
Che, W.; Wang, M.; Manning, C.; and Ting, L.
\newblock 2013.
\newblock Named entity recognition with bilingual constraints.
\newblock In {\em In Proceedings of the 2013 Conference of the North American
  Chapter of the Association for Computational Linguistics: Human Language
  Technologies},  52--62.
\newblock Atlanta, Georgia: Association for Computational Linguistics.

\bibitem[\protect\citeauthoryear{Chiu and Nichols}{2016}]{chiu2016}
Chiu, J.~P., and Nichols, E.
\newblock 2016.
\newblock Named entity recognition with bidirectional {LSTM-CNNs}.
\newblock {\em Transactions of the Association for Computational Linguistics}
  4:357--370.

\bibitem[\protect\citeauthoryear{Collobert \bgroup et al\mbox.\egroup
  }{2011}]{Collobert}
Collobert, R.; Weston, J.; Bottou, L.; Karlen, M.; Kavukcuoglu, K.; and Kuksa,
  P.
\newblock 2011.
\newblock Natural language processing (almost) from scratch.
\newblock {\em The Journal of Machine Learning Research}  2493--2537.

\bibitem[\protect\citeauthoryear{Dai and Cai}{2017}]{Dai}
Dai, F.~Z., and Cai, Z.
\newblock 2017.
\newblock Glyph-aware embedding of chinese characters.
\newblock \url{http://arxiv.org/abs/1709.00028}.
\newblock preprint.

\bibitem[\protect\citeauthoryear{Devlin \bgroup et al\mbox.\egroup
  }{2018}]{Devlin}
Devlin, J.; Chang, M.-W.; Lee, K.; and Toutanova, K.
\newblock 2018.
\newblock Bert: Pre-training of deep bidirectional transformers for language
  understanding.
\newblock \url{https://arxiv.org/abs/1810.04805}.
\newblock preprint.

\bibitem[\protect\citeauthoryear{Gers, Schmidhuber, and
  Cummins}{2000}]{Gers2000}
Gers, F.~A.; Schmidhuber, J.; and Cummins, F.
\newblock 2000.
\newblock Learning to forget: Continual prediction with {LSTM}.
\newblock {\em Neural Computation} 12(10):2451--2471.

\bibitem[\protect\citeauthoryear{Graves and Schmidhuber}{2005}]{Graves05}
Graves, A., and Schmidhuber, J.
\newblock 2005.
\newblock Framewise phoneme classification with bidirectional lstm and other
  neural network architectures.
\newblock {\em NEURAL NETWORKS}  5--6.

\bibitem[\protect\citeauthoryear{Hammerton}{2003}]{Hammerton2003}
Hammerton, J.
\newblock 2003.
\newblock Named entity recognition with long short-term memory.
\newblock In {\em Proceedings of the Seventh Conference on Natural Language
  Learning at HLT-NAACL 2003 - Volume 4}, CONLL '03,  172--175.
\newblock Stroudsburg, PA, USA: Association for Computational Linguistics.

\bibitem[\protect\citeauthoryear{He and Xu}{2017}]{oldweibo}
He, H., and Xu, S.
\newblock 2017.
\newblock A unified model for cross-domain and semi-supervised named entity
  recognition in {C}hinese social media.
\newblock In {\em AAAI Conference on Artificial Intelligence}.

\bibitem[\protect\citeauthoryear{Hochreiter and
  Schmidhuber}{1997}]{Hochreiter:1997:LSM:1246443.1246450}
Hochreiter, S., and Schmidhuber, J.
\newblock 1997.
\newblock Long short-term memory.
\newblock {\em Neural Comput.} 9(8):1735--1780.

\bibitem[\protect\citeauthoryear{Huang, Xu, and
  Yu}{2015}]{huang2015bidirectional}
Huang, Z.; Xu, W.; and Yu, K.
\newblock 2015.
\newblock Bidirectional {LSTM-CRF} models for sequence tagging.
\newblock {\em CoRR} abs/1508.01991.

\bibitem[\protect\citeauthoryear{Ioffe and Szegedy}{2015}]{batch}
Ioffe, S., and Szegedy, C.
\newblock 2015.
\newblock Batch normalization: accelerating deep network training by reducing
  internal covariate shift.
\newblock In {\em ICML'15 Proceedings of the 32nd International Conference on
  International Conference on Machine Learning}, volume~37,  448--456.

\bibitem[\protect\citeauthoryear{Li \bgroup et al\mbox.\egroup }{2015}]{Li}
Li, Y.; Li, W.; Sun, F.; and Li, S.
\newblock 2015.
\newblock Component-enhanced {C}hinese character embeddings.
\newblock \url{https://arXiv.org/abs/1508.06669}.
\newblock preprint.

\bibitem[\protect\citeauthoryear{Liu \bgroup et al\mbox.\egroup }{2017}]{Liu}
Liu, F.; Lu, H.; Lo, C.; and Neubig, G.
\newblock 2017.
\newblock Learning character-level compositionality with visual features.
\newblock \url{https://arxiv.org/abs/1704.04859}.
\newblock preprint.

\bibitem[\protect\citeauthoryear{Ma and Hovy}{2016}]{ma-hovy}
Ma, X., and Hovy, E.
\newblock 2016.
\newblock End-to-end sequence labeling via bi-directional {LSTM-CNNs-CRF}.
\newblock In {\em Proceedings of the 54th Annual Meeting of the Association for
  Computational Linguistics (Volume 1: Long Papers)},  1064--1074.
\newblock Berlin, Germany: Association for Computational Linguistics.

\bibitem[\protect\citeauthoryear{Meng \bgroup et al\mbox.\egroup
  }{2019}]{Glyce}
Meng, Y.; Wu, W.; Wang, F.; Li, X.; Nie, P.; Yin, F.; Li, M.; Han, Q.; Sun, X.;
  and Li, J.
\newblock 2019.
\newblock Glyce: Glyph-vectors for chinese character representations.
\newblock \url{http://arxiv.org/abs/1901.10125v3}.
\newblock preprint.

\bibitem[\protect\citeauthoryear{Nosirova, Xu, and Jiang}{2019}]{Nosirova}
Nosirova, N.; Xu, M.; and Jiang, H.
\newblock 2019.
\newblock A multi-task learning approach for named entity recognition using
  local detection.
\newblock https://arxiv.org/abs/1904.03300.

\bibitem[\protect\citeauthoryear{Pappu \bgroup et al\mbox.\egroup
  }{2017}]{pappu}
Pappu, A.; Blanco, R.; Mehdad, Y.; Stent, A.; and Kapil, T.
\newblock 2017.
\newblock Lightweight multilingual entity extraction and linking.
\newblock In {\em In Proceedings of the Tenth ACM International Conference on
  Web Search and Data Mining, WSDM ’17},  365--374.
\newblock New York, NY, USA: Association for Computational Linguistics.

\bibitem[\protect\citeauthoryear{Peng and Dredze}{2015}]{Weibo}
Peng, N., and Dredze, M.
\newblock 2015.
\newblock Named entity recognition for {C}hinese social media with jointly
  trained embeddings.
\newblock In {\em Empirical Methods in Natural Language Processing},  548--554.

\bibitem[\protect\citeauthoryear{Pradhan and
  Ramshaw}{2017}]{pradhan2017ontonotes}
Pradhan, S., and Ramshaw, L.
\newblock 2017.
\newblock Ontonotes: Large scale multi-layer, multi-lingual, distributed
  annotation.
\newblock In {\em Handbook of Linguistic Annotation}. Springer.
\newblock  521--554.

\bibitem[\protect\citeauthoryear{Pradhan \bgroup et al\mbox.\egroup
  }{2013}]{pradhan2013ontonotes}
Pradhan, S.; Moschitti, A.; Xue, N.; Ng, H.~T.; Bjorkelund, A.; Uryupina, O.;
  Zhang, Y.; and Zhong, Z.
\newblock 2013.
\newblock Towards robust linguistic analysis using ontonotes.
\newblock In {\em Proceedings of the Seventeenth Conference on Computational
  Natural Language Learning}.

\bibitem[\protect\citeauthoryear{Sang and Meulder}{2003}]{Gaz}
Sang, E. F. T.~K., and Meulder, F.~D.
\newblock 2003.
\newblock Introduction to the conll-2003 shared task: language-independent
  named entity recognition.
\newblock In {\em CONLL '03 Proceedings of the seventh conference on Natural
  language learning at HLT-NAACL 2003}, volume~4,  142--147.

\bibitem[\protect\citeauthoryear{Shi \bgroup et al\mbox.\egroup }{2015}]{Shi}
Shi, X.; Zhai, J.; Yang, X.; Xie, Z.; and Liu, C.
\newblock 2015.
\newblock Radical embedding: Delving deeper to {C}hinese radicals.
\newblock In {\em In Proceedings of the 53rd Annual Meeting of the Association
  for Computational Linguistics and the 7th International Joint Conference on
  Natural Language Processing (Volume 2: Short Papers)},  594--598.

\bibitem[\protect\citeauthoryear{Su and Lee}{2017}]{Su}
Su, T.-R., and Lee, H.-Y.
\newblock 2017.
\newblock Learning {C}hinese word representations from glyphs of characters.
\newblock \url{http://arxiv.org/abs/1708.04755}.
\newblock preprint.

\bibitem[\protect\citeauthoryear{Sun \bgroup et al\mbox.\egroup }{2014}]{Sun}
Sun, Y.; Lin, L.; Yang, N.; Ji, Z.; and Wang, X.
\newblock 2014.
\newblock Radical-enhanced {C}hinese character embedding.
\newblock In {\em In International Conference on Neural Information
  Processing},  279--286.
\newblock Springer.

\bibitem[\protect\citeauthoryear{Xu, Jiang, and Watcharawittayakul}{2017}]{xu}
Xu, M.; Jiang, H.; and Watcharawittayakul, S.
\newblock 2017.
\newblock A local detection approach for named entity recognition and mention
  detection.
\newblock In {\em In Proceedings of the 55th Annual Meeting of the Association
  for Computational Linguistics (Volume 1: Long Papers)},  1237--1247.
\newblock Vancouver, Canada: Association for Computational Linguistics.

\bibitem[\protect\citeauthoryear{Zhang and LeCun}{2017}]{Zhang}
Zhang, X., and LeCun, Y.
\newblock 2017.
\newblock Which encoding is the best for text classification in {C}hinese,
  {E}nglish, {J}apanese and {K}orean?
\newblock \url{https://arxiv.org/abs/1708.02657}.
\newblock preprint.

\bibitem[\protect\citeauthoryear{Zhang and Yang}{2018}]{latticeLSTM}
Zhang, Y., and Yang, J.
\newblock 2018.
\newblock {C}hinese {NER} using lattice {LSTM}.
\newblock In {\em Proceedings of the 56th Annual Meeting of the Association for
  Computational Linguistics (Volume 1: Long Papers)},  1554--1564.
\newblock Melbourne, Australia: Association for Computational Linguistics.

\bibitem[\protect\citeauthoryear{Zhou \bgroup et al\mbox.\egroup }{2015}]{AE}
Zhou, Y.; Arpit, D.; Nwogu, I.; and Govindaraju, V.
\newblock 2015.
\newblock Is joint training better for deep auto-encoders?
\newblock \url{https://arxiv.org/abs/1405.1380}.
\newblock preprint.

\bibitem[\protect\citeauthoryear{Zhu and Wang}{2019}]{zhu-wang-2019-ner}
Zhu, Y., and Wang, G.
\newblock 2019.
\newblock {CAN}-{NER}: {C}onvolutional {A}ttention {N}etwork for {C}hinese
  {N}amed {E}ntity {R}ecognition.
\newblock In {\em Proceedings of the 2019 Conference of the North {A}merican
  Chapter of the Association for Computational Linguistics: Human Language
  Technologies, Volume 1 (Long and Short Papers)},  3384--3393.
\newblock Minneapolis, Minnesota: Association for Computational Linguistics.

\end{thebibliography}
\bibliographystyle{aaai}
\end{document}